\documentclass[10pt,twocolumn,letterpaper]{article}

\usepackage{cvpr}
\usepackage{times}
\usepackage{epsfig}
\usepackage{graphicx}
\usepackage{amsmath}
\usepackage{amssymb}

\usepackage{color}

\usepackage{subfigure}

\usepackage{interval}

\usepackage{caption}

\usepackage{pifont}

\usepackage{makecell}
\usepackage{booktabs}
\usepackage{multirow}
\usepackage{siunitx}

\usepackage[T1]{fontenc}
\usepackage{fix-cm}

\newcolumntype{?}{!{\vrule width 1pt}}
\newcolumntype{C}[1]{>{\centering}m{#1}}

\newcolumntype{X}{@{\hskip\tabcolsep\vrule width 1.5pt\hskip\tabcolsep}}

\newcommand{\ours}{\textsc{DiMoFs}\xspace}
\newcommand{\our}{\textsc{DiMoFs}\xspace}

\usepackage{lipsum} 

\usepackage[pagebackref=true,breaklinks=true,letterpaper=true,colorlinks,bookmarks=false]{hyperref}

 \cvprfinalcopy 

\ifcvprfinal\fi
\begin{document}

\title{Learning Discriminative Motion Features Through Detection}

\author{Gedas Bertasius$^{1,2}$, Christoph Feichtenhofer$^{1}$, Du Tran$^{1}$, Jianbo Shi$^{2}$, Lorenzo Torresani$^{1,3}$\\
$^{1}$Facebook Research, $^{2}$University of Pennsylvania, $^{3}$Dartmouth College}


\maketitle

\begin{abstract}


Despite huge success in the image domain, modern detection models such as Faster R-CNN have not been used nearly as much for video analysis. This is arguably due to the fact that detection models are designed to operate on single frames and as a result do not have a mechanism for learning motion representations directly from video. We propose a learning procedure that allows detection models such as Faster R-CNN to learn motion features directly from the RGB video data while being optimized with respect to a pose estimation task. Given a pair of video frames---Frame A and Frame B---we force our model to predict human pose in Frame A using the features from Frame B. We do so by leveraging deformable convolutions across space and time. Our network learns to spatially sample features from Frame B in order to maximize pose detection accuracy in Frame A. This naturally encourages our network to learn motion offsets encoding the spatial correspondences between the two frames. We refer to these motion offsets as \ours ({\em Di}scriminative {\em Mo}tion {\em F}eature{\em s}).

In our experiments we show that our training scheme helps learn effective motion cues, which can be used to estimate and localize salient human motion. Furthermore, we demonstrate that as a byproduct, our model also learns features that lead to improved pose detection in still-images, and better keypoint tracking. Finally, we show how to leverage our learned model for the tasks of spatiotemporal action localization and fine-grained action recognition.

\end{abstract}
\vspace{-0.5cm}

\section{Introduction}

Modern CNN based detection models have been highly successful on various image understanding tasks such as edge detection, object detection, semantic segmentation, and pose detection~\cite{SPP,he2017maskrcnn,lin2017focal,ren2015faster,girshick15fastrcnn,girshick2014rcnn,guptaECCV14,44872,dai16rfcn,DBLP:conf/cvpr/RedmonDGF16,DBLP:journals/corr/RedmonF16,gberta_2017_CVPR,crfasrnn_iccv2015,cao2017realtime,wei2016cpm,gberta_2015_CVPR,DBLP:journals/corr/XieT15}. However, such models lack the means to learn motion cues from video data, and thus, they have not been used as widely for video understanding tasks where motion information plays a critical role.




\captionsetup{labelformat=default}
\captionsetup[figure]{skip=10pt}

\begin{figure}
\begin{center}
   \includegraphics[width=0.95\linewidth]{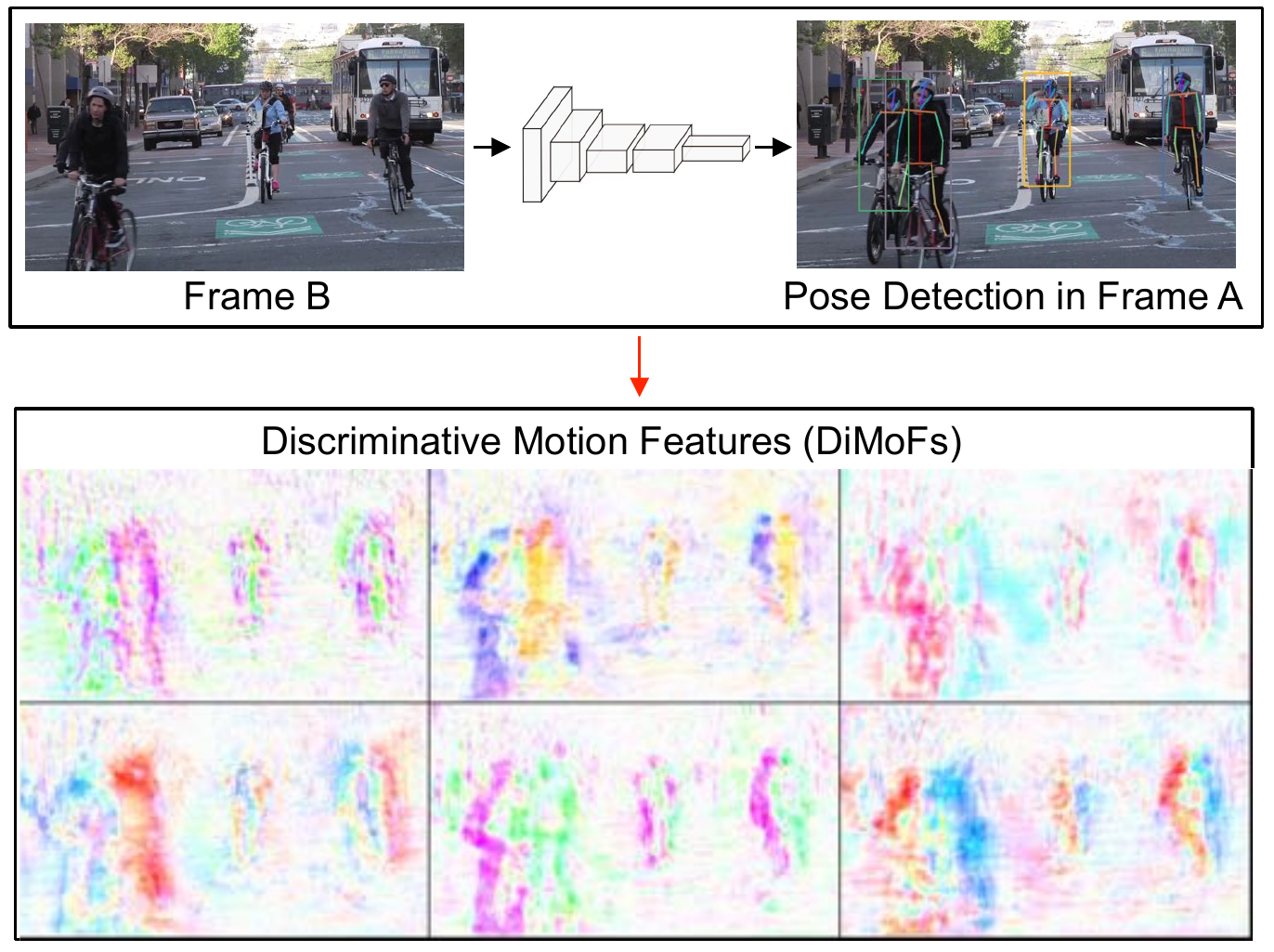}
\end{center}
\vspace{-0.5cm}
   \caption{We extend modern detection models (e.g., Faster R-CNN) with the ability to learn discriminative motion features (\ours) from RGB video data. We do so by training the detector on pairs of time-separated frames, with the objective of predicting pose in one frame using features from the other frame. This task forces the network to learn  motion ``offsets'' relating the two frames but that are discriminatively optimized for detection. Our learned \ours can be used on a variety of applications: salient motion localization, human motion estimation, improved pose detection and keypoint tracking, spatiotemporal action localization, and fine-grained action recognition.  \vspace{-0.5cm}}
   
\label{main_fig}
\end{figure}

Prior work~\cite{peng:hal-01349107,NIPS2014_5353,Feichtenhofer16} has attempted to address this issue by adopting two-stream architectures, where one stream learns appearance features, while the other stream aims to learn motion cues extracted from optical flow inputs. However, these models are expensive since they require processing optical flow with an additional CNN stream.


Recently, there have been many attempts to learn effective video representations with 3D convolutional networks~\cite{Tran:2015:LSF:2919332.2919929, DBLP:conf/cvpr/CarreiraZ17}. In principle, such models are capable of capturing salient temporal and motion features, directly optimized for the end task. However, due to their high computational cost, 3D CNNs typically operate on very small spatial resolution inputs, and as a result are not very good at detecting fine grained visual cues such as the pose of a person. Furthermore, it has been recently shown that 3D CNNs optimized from RGB data for action recognition tend to be surprisingly insensitive to temporal ordering of frames~\cite{Huang2018WhatMA}. This suggests that in practice they capture little motion information. Thus, it is typically necessary to apply them to optical flow inputs in order to effectively leverage motion.

In this work, we propose a learning scheme that allows modern detection models (e.g., Faster R-CNN) to learn motion cues directly from the RGB video data while being optimized for a discriminative video pose estimation task. Given a pair of annotated frames from the same video---Frame A and Frame B---we train our model to detect pose in Frame A, using the features from Frame B. To achieve this goal, our model leverages deformable convolutions~~\cite{8237351} across space and time. Through this mechanism, our model learns to sample features from Frame B that maximize pose detection accuracy in Frame A. As a byproduct of our optimization, our model also  learns ``offsets'' capturing the motion relating Frame A to Frame B. We refer to these offsets as {\em \ours} ({\bf Di}scriminative {\bf Mo}tion {\bf F}eature{\bf s})  to emphasize that they are discriminatively optimized for detection. In our experiments, we show that our \our model pretrained on pose estimation can subsequently be used for salient motion localization, human motion estimation, improved pose detection, keypoint tracking, spatiotemporal action localization, and fine-grained action recognition.


\section{Related Work}


\textbf{Detection in Images.} Modern object detectors~\cite{SPP,he2017maskrcnn,lin2017focal,ren2015faster,girshick15fastrcnn,girshick2014rcnn,guptaECCV14,44872,dai16rfcn,DBLP:conf/cvpr/RedmonDGF16,DBLP:journals/corr/RedmonF16} are built using deep CNNs~\cite{NIPS2012_4824,Simonyan14c,He2016DeepRL}. One of the earlier of such object detection systems was R-CNN~\cite{girshick2014rcnn}, which operated in a two-stage pipeline, first extracting object proposals, and then classifying each of them using a CNN. To reduce the computational cost, RoI pooling operation was introduced in~\cite{SPP,girshick15fastrcnn}. A few years ago, Faster R-CNN~\cite{ren2015faster} replaced region proposal methods by another network, thus eliminating a two stage pipeline. Several methods~\cite{DBLP:conf/cvpr/RedmonDGF16,DBLP:journals/corr/RedmonF16} extended Faster R-CNN into a system that runs in real time with little loss in performance. The recent Mask R-CNN~\cite{he2017maskrcnn} introduced an extra branch that predicts a mask for each region of interest, Finally, Deformable CNNs~\cite{8237351} leveraged deformable convolution to model deformations of objects more robustly. While these prior detection methods work well on images, they are not designed to exploit motion cues in a video-- a shortcoming we aim to address.

\textbf{Detection in Videos.} Several recent methods proposed architectures capable of aligning features temporally for improved object detection in video~\cite{zhu17fgfa,DBLP:conf/eccv/BertasiusTS18,xiao-eccv2018}. The method in~\cite{xiao-eccv2018} proposes a spatial-temporal memory mechanism, whereas~\cite{DBLP:conf/eccv/BertasiusTS18} leverages spatiotemporal sampling for feature alignment. Furthermore, the work in~\cite{zhu17fgfa} uses an optical flow CNN to align the features across time. 

While the mechanisms in~\cite{DBLP:conf/eccv/BertasiusTS18,xiao-eccv2018} are useful for improved detection, it is not clear how to use them for motion cue extraction, which is our primary objective. Furthermore, models like~\cite{zhu17fgfa} are redundant since they compute flow for every single pixel, which is rarely necessary for higher level video understanding tasks. Using optical flow CNN also adds $40M$ extra parameters to the model, which is costly.


\textbf{Two-Stream CNNs.} Recently, two-stream CNN architectures have been a popular choice for incorporating motion cues into modern CNNs~\cite{peng:hal-01349107,NIPS2014_5353,Bilen16a,10.1109/TPAMI.2017.2769085,Girdhar_17b_AttentionalPoolingAction,Feichtenhofer16,feichtenhofer2016spatiotemporal,feichtenhofer2017multiplier}. In these types of models, one stream learns appearance features from RGB data, whereas the other stream learns motion representation from the manually extracted optical flow inputs. The work in~\cite{Bilen16a, 10.1109/TPAMI.2017.2769085} leverages two-stream architectures for learning more effective spatiotemporal representations.  Recent methods in~\cite{Wang_towardsgood, conf/cvpr/NgHVVMT15,feichtenhofer2016spatiotemporal} explored the use of different backbone networks for action recognition tasks. Furthermore, various techniques have explored how to fuse the information from two streams~\cite{Feichtenhofer16,feichtenhofer2017multiplier,Girdhar_17b_AttentionalPoolingAction}. However, these two-stream CNNs are costly and consume lots of memory. Learning discriminative motion cues from the RGB video data directly instead of relying on manually extracted optical flow inputs could alleviate this issue.




\begin{figure*}
\begin{center}
   \includegraphics[width=0.9\linewidth]{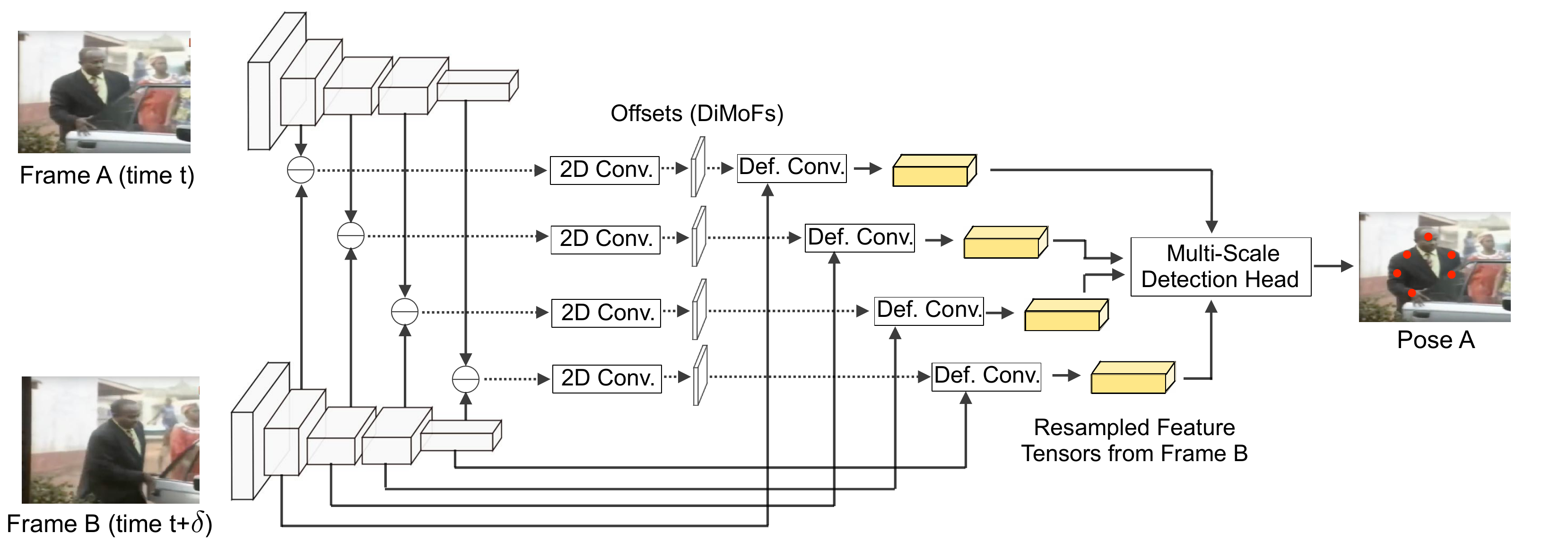}
\end{center}
\vspace{-0.5cm}
\caption{An illustration of our Discriminative Motion Feature (\our) training procedure. Given Frame A and Frame B, which are separated by $\delta$ steps in time, our goal is to detect pose in Frame A using the features from Frame B. First, we extract multi-scale features from both frames via a backbone CNN with shared parameters. Then, at each scale, we compute the difference between feature tensors A and B. From these tensor differences, offsets $\Delta p_n$ are predicted for each pixel location $p_n$. The predicted offsets are used to re-sample feature tensor B. As a last step, the resampled feature tensors from each scale are fed into a multi-scale detection head, which is used to predict the pose in Frame A. Our scheme optimizes end-to-end the \our network so that the feature tensors re-sampled from Frame B maximize the pose detection accuracy in Frame A. \vspace{-0.4cm}}
\label{arch_detection_fig}
\end{figure*}

\textbf{3D CNNs.} Currently the most common approach for learning features from raw RGB videos is via 3D convolutional networks~\cite{Tran:2015:LSF:2919332.2919929}. Whereas the method in~\cite{Tran:2015:LSF:2919332.2919929} proposes a 3D network architecture for end-to-end feature learning, the recent I3D method~\cite{DBLP:conf/cvpr/CarreiraZ17} inflates all 2D filters to 3D, which allows re-using the features learned in the image domain. Additionally, there have been many recent attempts at making 3D CNNs more effective by replacing 3D convolution with separable 2D and 1D convolutions~\cite{ DBLP:conf/eccv/XieSHTM18,DBLP:journals/corr/abs-1711-11248,Chollet2017XceptionDL,qiu2017learning}.  

While modern 3D CNNs are effective on popular action recognition datasets~\cite{DBLP:conf/cvpr/CarreiraZ17, Soomro_ucf101:a, Kuehne11},  3D CNNs are not designed for tasks that require detection of fine-grained visual cues since they operate on small spatial resolution to accommodate long video clips. Furthermore, it has been shown that 3D CNNs do not actually learn motion cues~\cite{Huang2018WhatMA}, and that they still need to rely on two stream architectures. 


\textbf{Relational Reasoning.} There have been several methods modeling temporal relations in videos~\cite{Wang_Transformation, zhou2017temporalrelation}. The work in~\cite{Wang_Transformation} learns weights for modeling "cause and effect" type of relationship. Furthermore, a similar method to ours~\cite{zhou2017temporalrelation} proposes a temporal relational module  for reasoning about temporal dependencies between video frames. However, the proposed relational module does not actually learn motion cues, and works effectively only when videos have strong temporal order~\cite{zhou2017temporalrelation}, which many datasets do not~\cite{DBLP:conf/cvpr/CarreiraZ17, Soomro_ucf101:a, Kuehne11}. In contrast, we aim to learn discriminative motion cues, which encode more general information and should be useful even if videos are not temporally ordered.



\section{Learning Discriminative Motion Features}
\label{detset_sec}

Our goal is to define a training procedure to learn discriminative motion features (\ours) from RGB videos using a detection model (e.g., Faster R-CNN). Consider the example in Figure~\ref{arch_detection_fig}, which shows a man closing the car door. In order to recognize this action (i.e., closing the door), we do not need to know the motion of every single pixel in the image. Just knowing how the hand of the person moves relative to the car gives us enough information about the action. The key question is: how can we learn such motion information discriminatively?  To do this, we formulate the following task. Given two video frames---Frame A and Frame B---our model is allowed to compare Frame A to Frame B but it must predict Pose A (i.e., the pose in Frame A) using the features from Frame B.

At first glance, this task may look overly challenging: how can we predict Pose A  by merely using features from Frame B? However, suppose that we had body joint correspondences between Frame A and Frame B. In such a scenario, this task would become trivial, as we would simply need to spatially ``warp'' the feature maps computed from frame B according to the set of correspondences relating frame B to frame A. Based on this intuition, we design a learning procedure that achieves two goals: 1) By comparing Frame A and Frame B, our model must be able to extract motion offsets relating these two frames. 2) Using these motion offsets our model must be able to rewarp the feature maps extracted from Frame B in order to optimize the accuracy of pose detection in Frame A.

To achieve these goals, we first feed both frames through a backbone CNN with shared parameters. The aim of the backbone is to extract high-level discriminative features facilitating the computation of pose correspondences from the two frames. Then, the backbone feature maps from both frames are used to determine which feature locations from Frame B should be sampled for detection in Frame A. Finally, the resampled feature tensor from Frame B is used as input to the pose detection subnetwork which is optimized to maximize accuracy of Pose A. 



\textbf{Backbone Architecture.} As our backbone network we use a Feature Pyramid Network~\cite{lin2016fpn} based on a ResNet-101~\cite{He2016DeepRL} design. We note, however, that our system is not constrained to this specific architecture design, and that it can easily integrate other backbone architectures as well.


%


\begin{figure*}[t]
\begin{center}
   \includegraphics[width=0.95\linewidth]{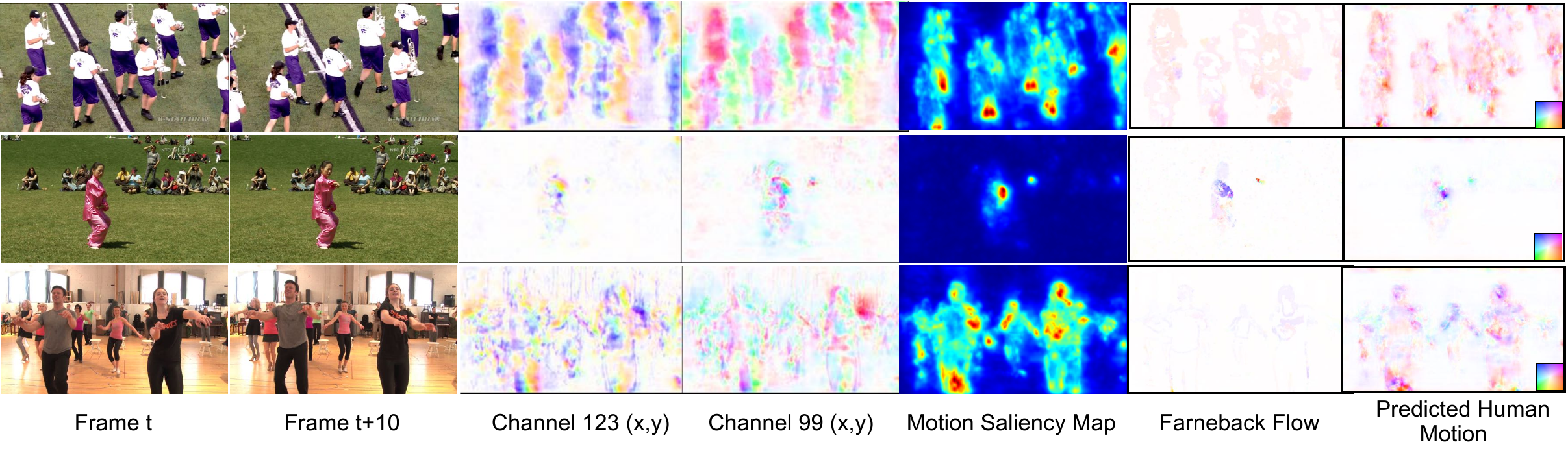}
\end{center}
\vspace{-0.7cm}
\caption{A figure with our \ours visualizations. In the first two columns, we visualize a pair of video frames that are used as input by our model. The $3^{rd}$ and $4^{th}$ columns depict $2$ (out of $256$) randomly selected \our channels visualized as a motion field. It can be noticed that the \ours capture primarily human motion, as they have been optimized for pose detection. Different channels appear to capture the motion of different body parts, thus performing a sort of motion decomposition of discriminative regions in the video. In the $5^{th}$ column, we display the magnitudes of summed \ours channels, which highlight salient human motion. Finally, the last two columns illustrate the standard Farneback flow, and the human motion predicted from our \ours. To predict human motion we train a \textbf{linear} classifier to regress the $(x,y)$ displacement of each joint from the offset maps. The color wheel, at the bottom right corner encodes motion direction.\vspace{-0.4cm}}
\label{sal_motion_wOF_fig}
\end{figure*}
\captionsetup{labelformat=default}
\captionsetup[figure]{skip=10pt}


\textbf{Learning to Sample Features.} Initially, we feed Frame A and Frame B through our backbone CNN, which outputs feature tensors \smash{$f^{(s)}_A$} and \smash{$f^{(s)}_B$} at four scales $s\in\{1,2,3,4\}$. Then, we compute the difference \smash{$d^{(s)}_{A,B} = f^{(s)}_A - f^{(s)}_B$}, at each of the four scales. The resulting feature tensor \smash{$d^{(s)}_{A,B}$} is  provided as input to a 2D convolutional layer, which predicts offsets $\Delta p_n$ (\ours) at all locations $p_n$. The offsets are used to spatially rewarp (sample) the feature tensor $f^{(s)}_{B}$. 

We implement the sampling mechanism via a deformable convolutional layer~\cite{8237351}, which takes 1) the predicted offsets $\Delta p_n$, and 2) the feature tensor $f^{(s)}_{B}$ as its inputs, and then outputs a newly sampled feature tensor \smash{$g^{(s)}_{A,B}$}.  Intuitively, the newly resampled feature tensor $g^{(s)}_{A,B}$ should encode all the relevant information needed for accurate pose detection in Frame A. We use subscript $(A,B)$ for \smash{$g^{(s)}_{A,B}$} to indicate that even though  \smash{$g^{(s)}_{A,B}$} was resampled from tensor \smash{$f^{(s)}_{B}$}, the offsets for resampling were computed using \smash{$d^{(s)}_{A,B}$}, which contains information about both feature tensors. An illustration of our architecture is presented in Figure~\ref{arch_detection_fig}.

\textbf{Multi-Scale Detection Head.} We employ a multi-scale detection head as is done in the Feature Pyramid Network~\cite{lin2016fpn}. The outputs of RoI Align~\cite{he2017maskrcnn} applied on the resampled feature tensor $g^{(s)}_{A,B}$ are then fed into three branches optimized for the following 3 tasks: 1) binary classification (person or not person), 2) bounding box regression (predicting region bounds) and 3) pose estimation (outputting a probability heatmap per joint). Our classification and bounding box regression branches are based on the original Faster R-CNN design~\cite{ren2015faster}, whereas our pose heatmap branch is designed according to the architecture of the ICCV17 PoseTrack challenge winner~\cite{girdhar2018detecttrack}.

\section{Interpreting DiMoFs}

Before discussing how DiMoFs can be employed in discriminative tasks (Section~\ref{sec:applications}), we would like to first understand how the motion cues are encoded in \ours. 
It turns out that interpreting what the \ours have learned, is nearly as difficult as analyzing any other CNN features~\cite{journals/corr/ZeilerF13, journals/corr/YosinskiCNFL15}. The main challenge comes from the high dimensionality of \ours: we are predicting $256$ $(x,y)$ displacements for every pixel at each of the four scales.

In Columns $3,4$ of Figure~\ref{sal_motion_wOF_fig}, we visualize two randomly selected offset channels as a motion field. The \ours visualized here were obtained by training our model for pose estimation on the PoseTrack dataset, as further discussed in the experiments section. Based on this figure, it is clear that \ours focus on people in the videos. However, it is quite difficult to tell what kind of motion cues each of these channels is encoding. Specifically, different \our maps seem to encode different motions rather than all predicting the same solution (say, the optical flow between the two frames). This makes sense, as our \ours are discriminatively trained. The network may decide to ignore motions of uninformative regions, while different \our maps may capture the motion of different human body parts (say, a hand as opposed to the head). Thus, our \ours can be interpreted as producing a motion decomposition in different channels of the most discriminative cues in the video.

\subsection{Human Motion Localization and Estimation}
\label{motion_pred_sec}

First, we observe that the magnitudes of our learned \ours encode salient human motion, which we visualize in Column 5 of Figure~\ref{sal_motion_wOF_fig}. Then, to verify that our learned \ours encode human motion relating the two frames we propose a simple visualization. For each point $p_n$ denoting a body joint, we extract $2048$-dimensional ($512$ channels concatenated across $4$ scales) \our feature vector $f_n$ at that specific point. Then, a {\em linear} classifier is trained to regress the ground truth $(x,y)$ motion displacement of this body joint from the feature vector $f_n$. During inference, we apply our trained classifier on every pixel of the image in a fully convolutional manner via a $1\times1$ convolution.


In Column 7 of Figure~\ref{sal_motion_wOF_fig}, we visualize our predicted motion outputs. We show Farneback's optical flow in Column 6 of the same Figure. Note that our predicted human motion matches quite well Farneback optical flow, especially in regions containing people. The fact that we can recover flow with our very simple linear prediction scheme suggests that our learned \our capture cues related to human motion. Furthermore, we point out that compared to the standard Farneback optical flow, our motion fields look less noisy. 

\begin{figure}[t]
\begin{center}
   \includegraphics[width=0.86\linewidth]{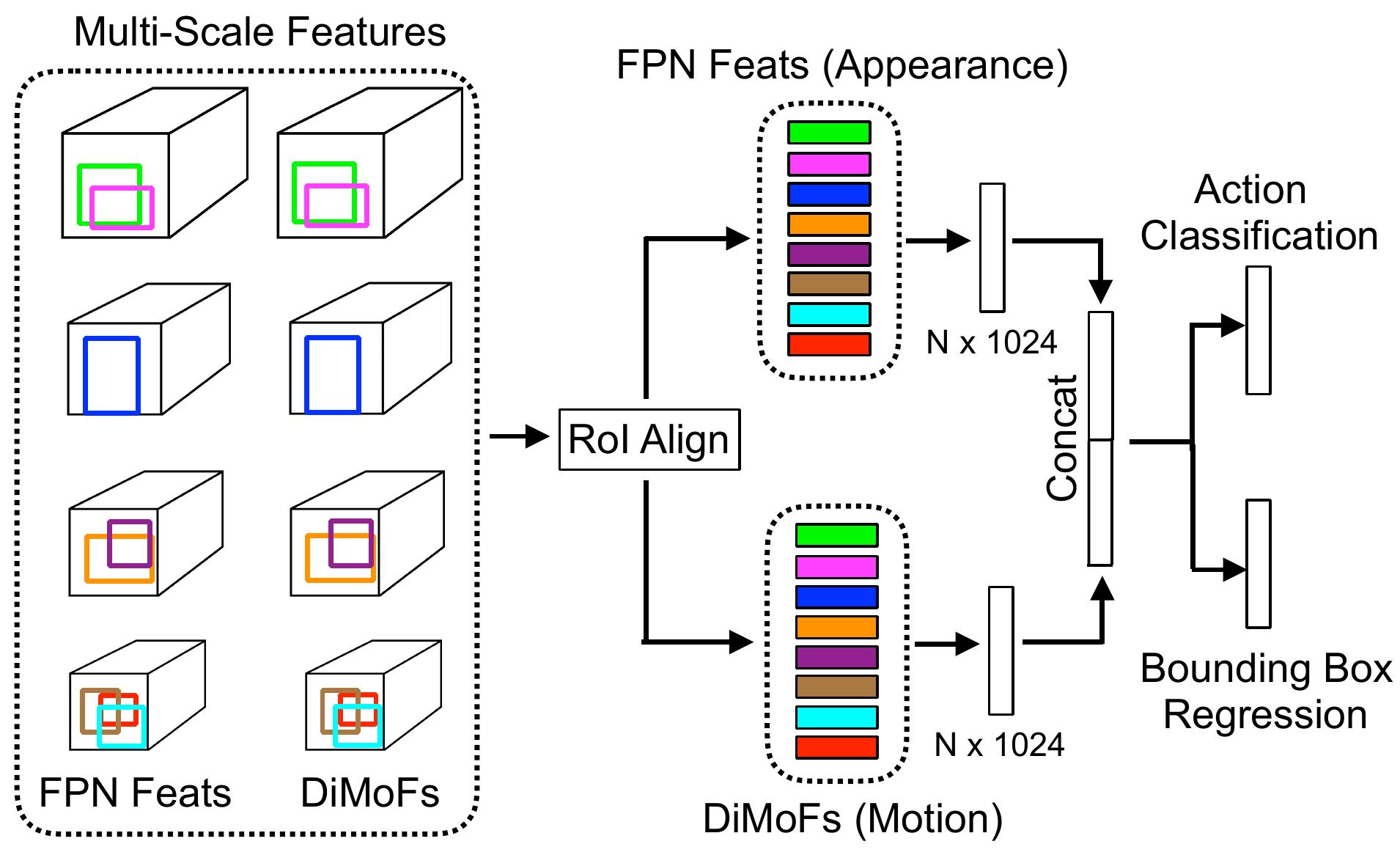}
\end{center}
\vspace{-0.6cm}
\caption{We extend our \our architecture for spatiotemporal action localization. Given a pair of video frames--Frame A and Frame B---we output for each person detected in Frame A, a bounding box and an action class. Up until the RoI Align, our model operates in the same way as for the pose detection task. Then, RoI Align is applied on 1) the multi-scale FPN feature tensors from Frame A (appearance), and 2) the multi-scale \ours (motion). These RoI features are then fed into separate MLPs, and the resulting $1024$-dimensional features are used to predict a bounding box and an action class for each RoI in Frame A.\vspace{-0.4cm}}
\label{arch_action_fig}
\end{figure}
\captionsetup{labelformat=default}
\captionsetup[figure]{skip=10pt}

\section{Detection and Recognition with DiMoFs}
\label{sec:applications}


In this section, we discuss how \ours benefit pose estimation and tracking. We also show how to adapt the DiMoFs architecture for the tasks of spatiotemporal action localization and fine-grained action recognition.



\textbf{Pose Detection.} During \textit{training} for a video pose detection task (see Section~\ref{detset_sec}), we select Frame B by sampling a frame $\delta$ time-steps from Frame A, where $\delta$ is randomly picked for each training frame from the set $\{-10, \hdots, 10\}$. During \textit{inference}, we similarly feed a pair of frames into our model, and output pose predictions for Frame A by using the features from Frame B. Note that this also includes a special case when $\delta=0$, meaning that Frame B is chosen to be the same as Frame A. In this special case, we can simply initialize the feature difference tensors  $d^{(s)}_{A,B}$ to zeros, and then proceed as before. This allows us to train our pose detector on videos, and then later test it on still-images. 


\textbf{Keypoint Tracking.} We consider the problem of tracking human body keypoints in video. We approach this task using the same ICCV17 PoseTrack winning model~\cite{girdhar2018detecttrack} as we did for pose detection. The tracking system in~\cite{girdhar2018detecttrack} consists of two stages: 1) keypoint estimation in frames, and 2) bipartite graph matching to link the tracks across adjacent frames. We simply replace the baseline model in stage 1) with our \our learned model. Note that both networks have the same network architecture, but differ in their training procedures. As will be shown later, our \our training scheme leads to better keypoint tracking results.

 \begin{figure}
\begin{center}
   \includegraphics[width=0.88\linewidth]{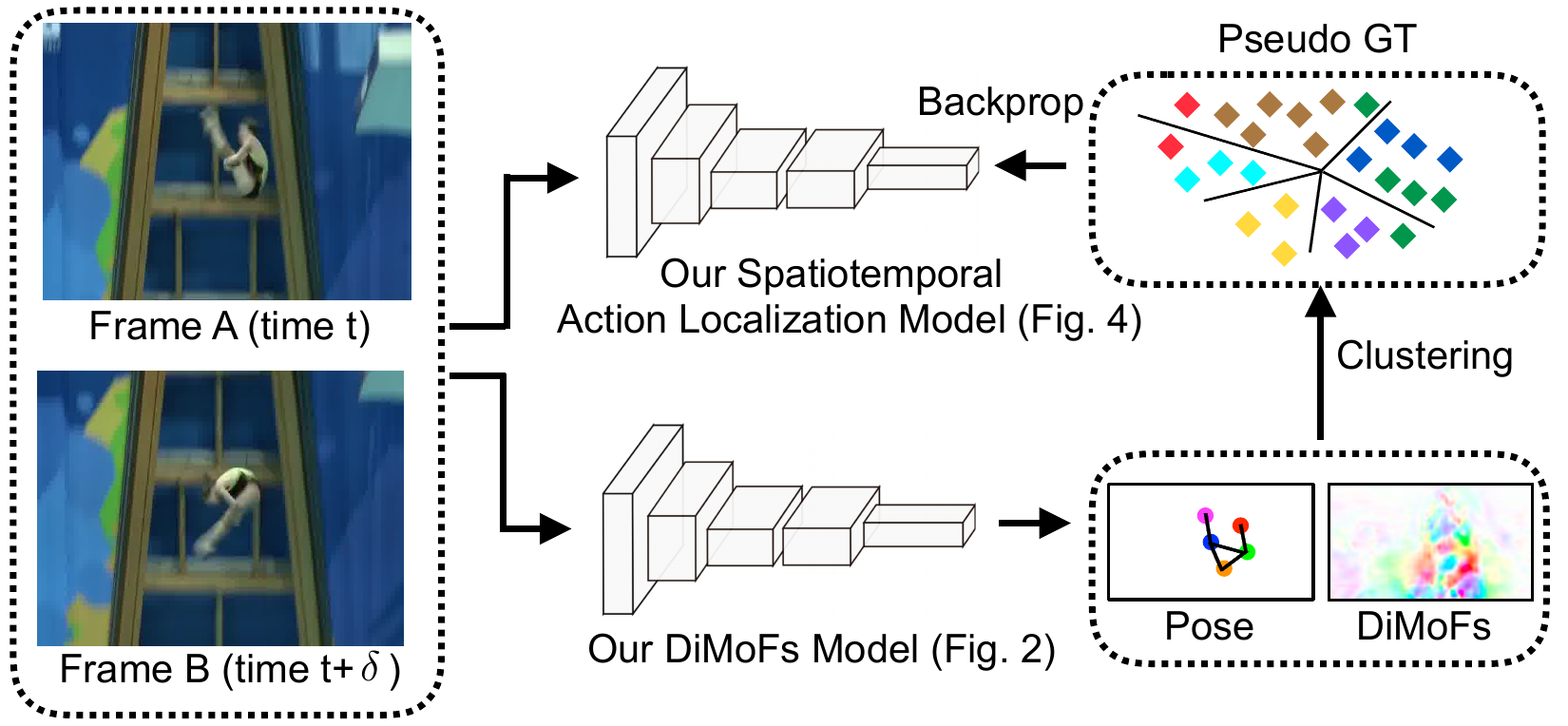}
\end{center}
\vspace{-0.4cm}
   \caption{ A high-level overview of our approach for learning fine-grained action recognition features. Initially, we use our \our model to extract pose and \ours features from a given pair of video frames. We then accumulate these pose and \our features across the entire Diving48 training set, and cluster them using a k-means algorithm. The resulting cluster assignment IDs are then used as pseudo ground truth labels to optimize our action recognition features, as illustrated in Figure~\ref{arch_action_fig}.\vspace{-0.4cm}}
\label{self_supervised_arch_fig}
\end{figure}

\subsection{Spatiotemporal Action Localization}
\label{action_loc_sec}

We introduce a few simple modifications to incorporate our learned \ours into the Faster R-CNN architecture for spatiotemporal action localization (see Figure~\ref{arch_action_fig}). Just as before, the model takes a pair of video frames as input. However, in this case the model outputs bounding boxes for Frame A, and an action class for each bounding box. We conjecture that a good spatiotemporal action localization model needs 1) strong visual appearance features associated with Frame A, and 2) motion features that encode how the person is moving between Frames A and B. While the appearance information is provided by the Frame A features computed through the backbone network, the motion cues are captured by our learned \ours. 


We keep the operations in the backbone the same as they were for the pose detection task. To adapt our pose estimation architecture to the problem of action localization, we remove all deformable convolutional layers as we do not need to resample features from B into Frame A anymore. Afterwards, we predict regions of interest (RoIs) in the same manner as we did for the pose detection task and then apply RoI Align separately on the visual features, and on the \ours. Afterwards, separate 2-layer MLP head is applied to each type of features, and the resulting feature tensors are concatenated together to predict the bounding box, as well as its associated action class (See Figure~\ref{arch_action_fig}).

\subsection{Fine-Grained Action Recognition}
\label{fine_grained_sec}

Our model operates on frame pairs and it is not designed to process long video sequences, unlike 3D CNNs for action recognition~\cite{Tran:2015:LSF:2919332.2919929, DBLP:conf/cvpr/CarreiraZ17, DBLP:conf/eccv/XieSHTM18,DBLP:journals/corr/abs-1711-11248,Chollet2017XceptionDL,qiu2017learning}. However, our \ours model is explicitly designed to detect fine-grained cues such as human pose and subtle movements of various body parts. This suggests that \ours can potentially be leveraged for fine-grained action recognition. We achieve this goal through a procedure inspired by the work of Caron et al.~\cite{DBLP:conf/eccv/CaronBJD18} who propose a two-stage deep clustering method for unsupervised learning of still-image features. Initially, we use our \our model (see Fig.~\ref{arch_detection_fig}) to extract the coordinates of each body part of every person in a video. We also extract \our features associated with each body joint and reduce them to $15$ dimensions using PCA. We then concatenate these features and cluster them via K-means using $K=100$ clusters. Intuitively, such a clustering procedure should yield clusters sharing similar pose and body motion patterns. We illustrate this procedure in Figure~\ref{self_supervised_arch_fig}.

\begin{figure}
\begin{center}
   \includegraphics[width=0.73\linewidth]{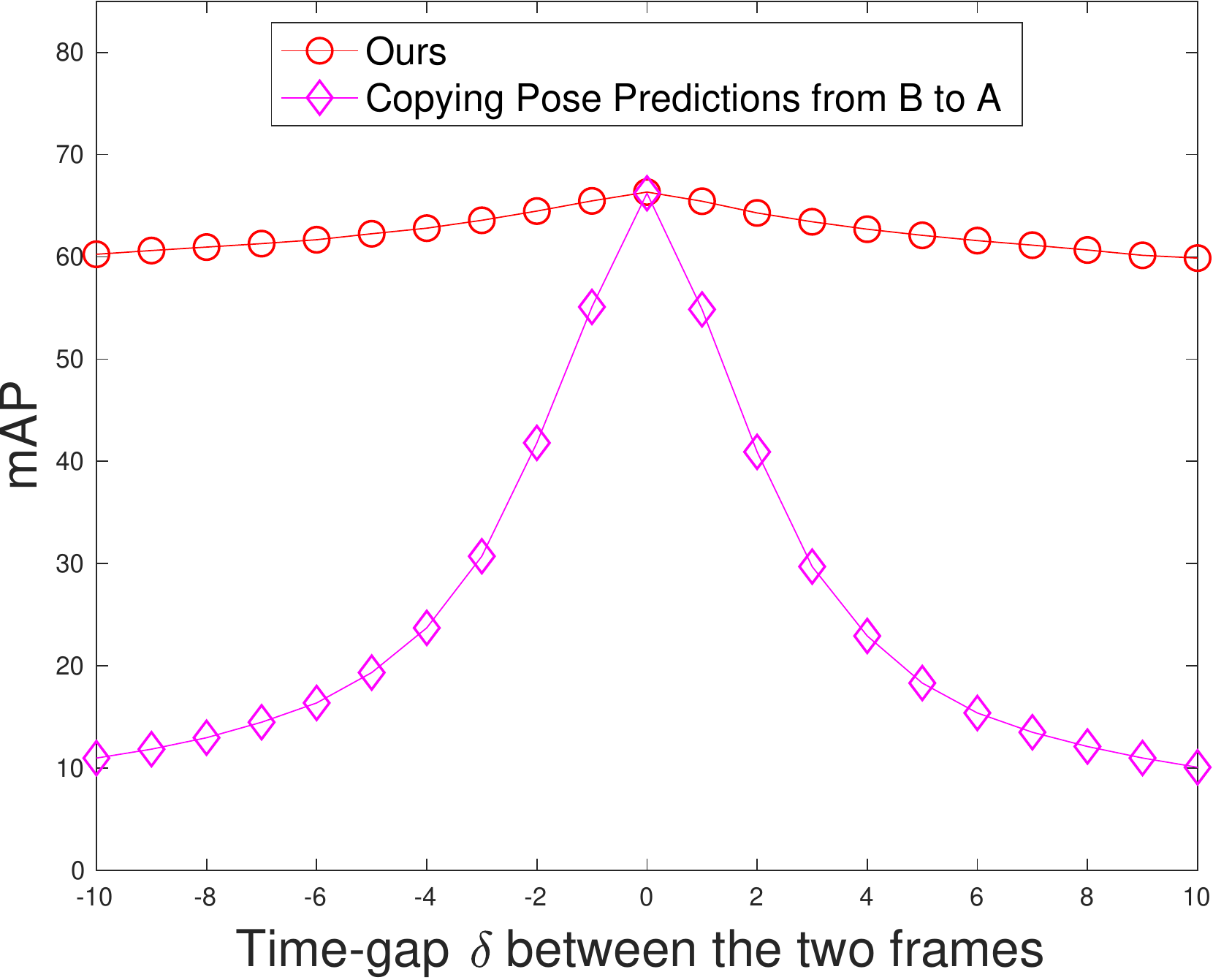}
\end{center}
\vspace{-0.4cm}
   \caption{Pose detection results on the PoseTrack dataset. Our model predicts pose in Frame A using features from a Frame B that is $\delta$ time-steps away.   During {\bf training}, our model is optimized on frame pairs with random time-gaps $\delta \in \{-10, \hdots,10\}$. Here, we present the results obtained by using different values of $\delta$ \textbf{at inference} time.
As $\delta$ becomes large the motion between frames may become substantial. Yet, the accuracy of our model drops gracefully as $\delta$ deviates from 0. This confirms that our model estimates the human motion between the two frames reliably.\vspace{-0.4cm}}
\label{pose_results_fig}
\end{figure} 

Afterwards, we train our spatiotemporal action localization network (see Figure~\ref{arch_action_fig}) to predict such cluster IDs. Thus, instead of predicting action classes, we use cluster IDs as pseudo-ground truth labels. The rationale behind this choice is that directly predicting action classes from pairs of frames is a nearly-impossible task, especially in the case of fine-grained classification where the same poses occur in many different categories. On the other hand, predicting pose cluster IDs is a better posed task. At the same time, the aggregation of pose cluster IDs over the entire video  (e.g. in the form of a simple histogram) can provide a strong descriptor for action recognition, disambiguating the recognition problem we had in a single pair of frames. Inspired by this intuition, we initialize our spatiotemporal action localization with weights learned during our \our training procedure and optimize it for pose cluster ID prediction. Afterwards, we evaluate the model on all pairs of frames in the video (sampled with time-gap $\delta=5$) and obtain the $100$ dimensional pose cluster ID prediction for each pair. The final classification is performed by training a shallow $2$-layer MLP on the $100$-dimensional vector obtained by summing up the pose ID activations for all pairs in the same video. 









\section{Experiments}

In this section, we present results showing the benefit of our \ours on the tasks discussed in the previous section: 1) pose detection, 2) keypoint tracking, 3) spatiotemporal action localization, and 4) fine-grained action recognition. 

\subsection{Pose Detection and Tracking on PoseTrack}

In this section, we train and test our method on the PoseTrack~\cite{Iqbal_CVPR2017} dataset, which contains 514 videos, 300 for training, 50 for validation and 208 for testing. The dataset includes 23,000 frames with annotated poses. Our evaluations are performed on the validation set. We demonstrate the effectiveness of our \our procedure by showing improved results for both pose estimation and keypoint tracking with respect to an analogous model trained on individual frames and thus unable to use motion information. 


\begin{table}[t]
\setlength{\tabcolsep}{2.0pt}
\footnotesize
\begin{center}
 \begin{tabular}{|c | c  c  c  c  c  c  c | c |} 
 \hline
 Method & Head & Shou & Elbo &  Wri & Hip &  Knee & Ank & Mean\\
 \hline\hline
 Girdhar et al.~\cite{girdhar2018detecttrack} & 72.8 & 75.6 & 65.3 & 54.3 & 63.5 & 60.9 & 51.8 &  64.1\\
  \cite{girdhar2018detecttrack} + DiMoFs & \bf 75.2 & \bf 78.5 & \bf 66.8 & \bf 56.0 & \bf 66.7 & \bf 62.7 & \bf 54.0 & \bf 66.3\\
 \hline
\end{tabular}
\end{center}
\vspace{-0.4cm}
\caption{Pose detection results on the PoseTrack dataset measured in mAP. The first row reports the results of a single-frame baseline~\cite{girdhar2018detecttrack}, which won the ICCV17 PoseTrack challenge. In the second row, we present the results of our method, which is based on the same exact model as~\cite{girdhar2018detecttrack} but it is {\em trained} on pairs of frames with our \our training scheme (the {\em testing} is still done on individual frames). These results indicate consistent performance gains for each body joint, and a mean mAP improvement of $2.2\%$.\vspace{-0.5cm}}
\label{pose_results_table}
\end{table}






The Figure~\ref{pose_results_fig} illustrates our pose detection results according to the mAP metric for different values of the time-gap $\delta$ used at {\em testing} time. The $\delta$ values  on the x-axis of Figure~\ref{pose_results_fig} captures how far apart Frames A and B are from each other. As expected, the more $\delta$ deviates from $0$, the lower the accuracy becomes because the motion between two frames becomes more severe. We discuss below two key findings from the results we obtain. 

\textbf{Results when $\boldsymbol{\delta \neq 0}$.} We first observe that the accuracy of our model drops quite gracefully as the time-gap $\delta$ deviates from $0$. By comparison, note the poor performance of the magenta curve which depicts a na\"{\i}ve baseline that simply copies pose detections from B to A (i.e., without using Frame A at all). For $\delta=\pm10$, the accuracy of such baseline drops by $56.5\%$ mAP with respect to the single-frame baseline~\cite{girdhar2018detecttrack}. In contrast, our model drops by only $6.1\%$ mAP for the same $\delta=\pm10$. The ability to predict poses from far-away frames suggests that our model reliably estimates the motion between the two frames and warps accurately the features of frame B for detection in frame A.

\textbf{Results when $\boldsymbol{\delta=0}$.} In Table~\ref{pose_results_table}, we compare our model with the single frame baseline of Girdhar et al.~\cite{girdhar2018detecttrack}, which won the ICCV17 PoseTrack challenge. We note that our method is based on the same CNN architecture as~\cite{girdhar2018detecttrack}, but is instead {\em trained} on pairs of frames using our \our training procedure.  The {\em testing} when $\delta=0$, however, is done on individual frames for both methods.

We observe that \our outperforms this strong baseline for all joints and by $2.2\%$ on average. 
What are the reasons behind these gains? This happens primarily because our \our training procedure on pairs of frames enables a form of data augmentation. In a single-frame training setting, the model learns a pattern connecting Frame A to Pose A. However, in our \our training, for each Frame A the model is learning patterns connecting many different Frames B (as $\delta$ is varied) to Label A. Thus, our model leverages many more (frame, label) pairs, which is beneficial.



\begin{table}[t]
\setlength{\tabcolsep}{2.0pt}
\footnotesize
\begin{center}
 \begin{tabular}{|c | c  c  c  c  c  c  c | c |} 
 \hline
 Method & Head & Shou & Elbo &  Wri & Hip &  Knee & Ank & Mean \\ [0.5ex] 
 \hline\hline
  Girdhar et al.~\cite{girdhar2018detecttrack}  & 61.7 & 65.5 & 57.3 & 45.7 & 54.3 & 53.1 & 45.7 & 55.2\\
 \cite{girdhar2018detecttrack} + DiMoFs & \bf 62.9 & \bf 67.2 & \bf 57.4 & \bf 45.8 & \bf 55.5 & \bf 53.5 & \bf 46.5 & \bf 56.0\\
 \hline
\end{tabular}
\end{center}
\vspace{-0.5cm}
\caption{Keypoint tracking results on the PoseTrack dataset using Multi-Object Tracking Accuracy (MOTA) metric. As our baseline, we use the ICCV17 PoseTrack challenge winning method~\cite{girdhar2018detecttrack}. Our model is trained using the same architecture as~\cite{girdhar2018detecttrack}, but using our \our training scheme. 
\vspace{-0.75cm}}
\label{keyps_results_table}
\end{table}

\textbf{Keypoint Tracking.} In Table~\ref{keyps_results_table}, we also demonstrate that our \our training procedure is helpful for the keypoint tracking task on the same PoseTrack dataset. The results are measured using the standard Multi-Object Tracking Accuracy (MOTA). We observe that \ours yields gains in keypoint tracking  across all body joints although not as substantial as in the case of pose estimation. We note that both methods in Table~\ref{keyps_results_table} use the same CNN architecture, but our model is trained using \our procedure.







\subsection{Results of Action Localization on JHMDB}

Next, we evaluate the effectiveness of our spatiotemporal action localization model (described in subsection~\ref{action_loc_sec}). To do this, we use the JHMDB~\cite{Jhuang:ICCV:2013} dataset, which contains 928 temporally-trimmed clips representing 21 action classes, with bounding box annotations. We report our results on split $1$ using the standard frame-mAP metric with an intersection-over-union (IOU) threshold set to $0.5$.

\begin{figure}
\begin{center}
   \includegraphics[width=0.74\linewidth]{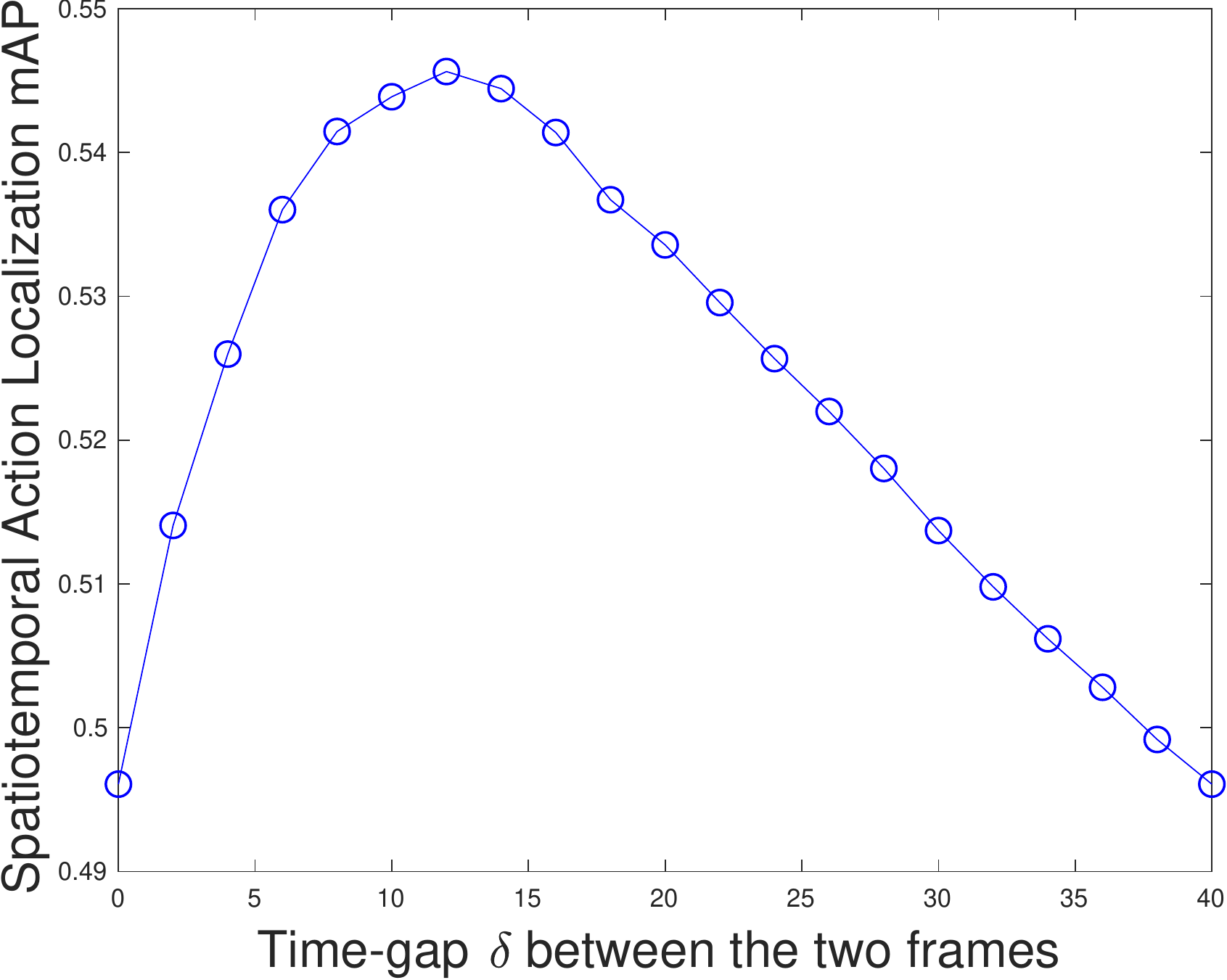}
\end{center}
\vspace{-0.5cm}
   \caption{Our spatiotemporal action localization results on JHMDB dataset as we vary the time-gap $\delta$ separating Frame B from Frame A during inference. Based on these results, we observe that the performance is lowest at $\delta=0$, which makes sense as there are no motion cues to leverage. However, once $\delta$ gets larger, the accuracy increases sharply indicating that the learned \ours contain useful motion cues for spatiotemporal action localization task.\vspace{-0.4cm}} 
\label{jhmdb_results_fig}
\end{figure}

To show that our new model can leverage the learned \ours for spatiotemporal action localization, we test our model by feeding it pairs of frames, where Frame B is sampled by choosing $\delta = \{0, 2, 4, 6, \hdots, 38, 40\}$. We illustrate these results in Figure~\ref{jhmdb_results_fig}, which displays spatiotemporal action localization mAP versus the time-gap $\delta$ between the two frames. From this figure, we observe that the accuracy is at its lowest when $\delta=0$, which makes sense because the model cannot leverage any motion cues.  However, once we increase $\delta$, the accuracy starts climbing steeply, which suggests that the model is leveraging the learned \our features for better spatiotemporal action localization performance. Based on the figure, we observe that the model reaches its peak performance at $\delta=12$, which corresponds to approximately $0.5$s in the original video. After that, the accuracy starts going down, which is expected because 1) the frames that are further away are less informative, and 2) it becomes harder to estimate long-range motion. 

Furthermore, in Table~\ref{jhmdb_results_table}, we ablate how different factors affect spatiotemporal action localization performance. First, we note that inflating a standard Faster R-CNN with the same backbone as our model to 3D~\cite{DBLP:conf/cvpr/CarreiraZ17} produces poor results (see column 1 of Table~\ref{jhmdb_results_table}). Next, in columns 2 and 3, we report the accuracy of the model that uses exactly the same model architecture but where \ours have not been trained for detection. From the table, we observe that such a model performs substantially worse ($44.3\%$ and $44.7\%$ for $\delta=0$ and $10$ respectively) than our \ours model ($49.6\%$ and  $54.4\%$ for $\delta=0$ and $10$ respectively). Furthermore, we point out that a model without \our training is not able to exploit motion cues effectively, which is indicated by a marginal $0.4\%$ improvement between $\delta=0$ and $\delta=10$ settings. In contrast, our model exhibits a substantial $4.8\%$ performance boost when increasing the $\delta$ from $0$ to $10$, which suggests the importance of learning \ours discriminatively through a detection task. It is also worth noting, that pose information learned via \our training is important, i.e., even in a setting where $\delta=0$, our \our model outperforms an equivalent Faster R-CNN by $5.3\%$  (see columns 2 and 4 in Table~\ref{jhmdb_results_table}).  Lastly, we point out that learning the appearance features as before, but replacing implicit motion cues encoded by \ours with explicit optical flow yields $50.6\%$ mAP at $\delta=10$, which is substantially worse than our $54.4\%$. 




We also note that while models pretrained using a Kinetics dataset yield better performance~\cite{gu2017,DBLP:conf/eccv/SunSVMSS18, DBLP:journals/corr/abs-1807-10066}, those results are not directly comparable to our evaluation since Kinetics is larger and contains many examples from the same classes as JHMDB. In this section, we aim to show that our model successfully leverages \ours to improve action localization without resorting to additional action labels.

 \setlength{\tabcolsep}{1pt}

 \begin{table}[t]
  \footnotesize
    \begin{center}
    \begin{tabular}{ l | c | c | c | c | c | c | c | }
    \hline
    
          &1 &2 & 3& 4 & 5 & 6\\ 
          \hline\hline
         \our Training? & &  & & \checkmark &  & \checkmark \\ \hline 
         Replacing \ours with Optical Flow? &  & & & & \checkmark  & \\ \hline
         Faster R-CNN Inflated to 3D?  & \checkmark & &  &  & & \\  \hline
         $\delta=0$ at Inference?  & & \checkmark & & \checkmark &  & \\ \hline
         $\delta=10$ at Inference?  & &  & \checkmark & & \checkmark  & \checkmark \\ \Xhline{3\arrayrulewidth}
	Mean Frame Average Precision (mAP) & 43.8 &  44.3 & 44.7  & 49.6  & 50.6  & \bf 54.4\\ \hline
    \end{tabular}
    \end{center}\vspace{-.4cm}
    \caption{We examine how various factors affect spatiotemporal action localization performance on JHMDB. We observe that learning \our discriminatively through detection outperforms the model with the same architecture but that was not trained with \our scheme by $5.3\%$ and $9.7\%$ mAP for $\delta=0$, and $\delta=10$ respectively (compare column $2$ to $4$, and $3$ to $6$). We also note that replacing  \ours offsets with optical flow yields $50.6\%$ mAP, which is $3.8\%$ worse than using \our (column $5$ vs $6$). \vspace{-0.4cm}}
    \label{jhmdb_results_table}
   \end{table}

\subsection{Fine-Grained Action Recognition on Diving48}

Finally, we evaluate our \ours model on a fine-grained action recognition task (see subsection~\ref{fine_grained_sec} for model description). For this experiment, we use the newly released Diving48 dataset~\cite{Li_2018_ECCV}, which contains $15,943$ training and $2096$ testing videos of professional divers performing $48$ types of dives. We choose this dataset because unlike datasets such as Kinetics or UCF101, Diving48 is designed to minimize the bias towards particular scenes or objects. To do well on Diving48 dataset, it is necessary to model the subtle differences in body motion cues, and use them to predict the action class (i.e., the type of dive). 

In Table~\ref{diving48_results_table}, we present our quantitative results. In the top half of the table, we examine performance of our model in comparison to recent 2D CNNs: TSN~\cite{TSN2016ECCV} and TRN~\cite{zhou2017temporalrelation}. We also include an interesting baseline that uses directly the histogram of cluster ID from pose detection (bottom branch in Fig.~\ref{self_supervised_arch_fig}) as features for action recognition (without training the top-branch in Fig.~\ref{self_supervised_arch_fig}). This 
yields a pretty solid accuracy of $17.2\%$. By including \ours (in addition to pose) for clustering, the accuracy jumps to $21.4\%$. Note, that if we replace implicit motion cues encoded by \ours with explicit optical flow the accuracy drops to $18.8\%$, which is substantially worse than our $21.4\%$. We also point out that training our spatiotemporal action recognition model to predict pose cluster IDs further improves the performance, and allows our model to achieve $24.1\%$. These results suggest 1) the usefulness of our learned pose and \our features, and also 2) highlight the benefits of optimizing the network to pseudo ground-truth pose cluster IDs.

 \setlength{\tabcolsep}{3pt}

 \begin{table}[t]
  \footnotesize
  \begin{center}
 \begin{tabular}{|c | c | c |} 
 \hline
 2D Models & Pre-training Data & Accuracy \\
 \hline\hline
 TSN (RGB)~\cite{TSN2016ECCV} & ImageNet (objects) & 16.8\\
 TSN (RGB + Flow)~\cite{TSN2016ECCV} & ImageNet  (objects) & 20.3\\
 TRN~\cite{zhou2017temporalrelation} & ImageNet  (objects) & 22.8\\ \hline
 Ours-init (Pose) & PoseTrack (poses) & 17.2\\
 Ours-init (Pose+Optical Flow) & PoseTrack (poses) & 18.8\\
 Ours-init (Pose+\ours) & PoseTrack (poses) & 21.4\\ \hline
 Ours-final (Pose+\ours) & PoseTrack (poses) & \bf 24.1  \\ \hline \addlinespace[2ex] \hline
 3D Models & Pre-training Data & Accuracy\\ 
 \hline\hline
 R(2+1)D~\cite{DBLP:journals/corr/abs-1711-11248} & None & 21.4 \\
 C3D~\cite{Tran:2015:LSF:2919332.2919929} & Sports1M (actions) & 27.6 \\
 R(2+1)D~\cite{DBLP:journals/corr/abs-1711-11248} & Kinetics (actions) & 28.9 \\
 Ours + R(2+1)D~\cite{DBLP:journals/corr/abs-1711-11248} & Kinetics + PoseTrack & \bf 31.4 \\ \hline
\end{tabular}
\end{center}\vspace{-0.4cm}
   \caption{Our results on Diving48 dataset. In the top half of the table, we show that our system achieves $24.1\%$ accuracy and outperforms 2D methods TSN~\cite{TSN2016ECCV} and TRN~\cite{zhou2017temporalrelation}. We also observe that even using the initial cluster ID histograms as features produces solid results (i.e. $17.2\%$ and $21.4\%$). Note that replacing \ours with optical flow reduces the accuracy by $2.6\%$ (compared to our $21.4\%$). Our model does not perform as well as the pre-trained R(2+1)D~\cite{DBLP:journals/corr/abs-1711-11248} baseline. However, our pre-training is done on a much smaller dataset, and on a more general pose detection task. Without Kinetics pre-training, we outperform  R(2+1)D~\cite{DBLP:journals/corr/abs-1711-11248} by a solid $2.7\%$ margin. Finally, we show that combining \ours with pre-trained R(2+1)D~\cite{DBLP:journals/corr/abs-1711-11248} improves the results, suggesting complementarity of the two methods. \vspace{-0.5cm}}
    \label{diving48_results_table}
   \end{table}
 
In the lower half of the table, we compare our method with popular 3D models~\cite{Tran:2015:LSF:2919332.2919929,DBLP:journals/corr/abs-1711-11248} . Our model does not perform as well as the pre-trained long-range 3D CNNs. However, this comparison is not exactly fair as these 3D models have been pre-trained on larger-scale action recognition dataset such as Kinetics~\cite{DBLP:conf/cvpr/CarreiraZ17}. In contrast, our model is pre-trained on a much smaller PoseTrack dataset, which requires significantly less computational resources. Furthermore, our pre-training is done on a more general pose estimation task, which allows our learned \ours to generalize to a variety of different tasks as shown in the paper. We note that without Kinetics pre-training, our model outperforms R(2+1)D~\cite{DBLP:journals/corr/abs-1711-11248} baseline by a solid  $2.7\%$ margin. Finally, we also show that combining our model with the pre-trained R(2+1)D~\cite{DBLP:journals/corr/abs-1711-11248} allows us to further improve the performance, and achieve state-of-the-art results, which suggests that the two models learn complementary cues.

\section{Conclusions}


In this work, we introduced \ours, discriminatively trained offsets that encode human motion cues. We showed that our \ours can be used  to localize and estimate salient human motion. Furthermore, we show that as a byproduct of our \our learning scheme, our model also learns features that lead to improved pose detection, and better keypoint tracking. Finally, we showed how to leverage our learned \our features for spatiotemporal action localization and fine-grained action recognition tasks.  Our future work involves designing unified CNN architectures that can leverage the strengths of modern 3D CNNs, and detection-based systems. We will release our source code and our trained models upon publication of the paper.




{\small
\bibliographystyle{ieee}
\bibliography{gb_bibliography}
}

\end{document}